\documentclass[sigconf]{aamas}  

\AtBeginDocument{%
  \providecommand\BibTeX{{%
    \normalfont B\kern-0.5em{\scshape i\kern-0.25em b}\kern-0.8em\TeX}}}

\usepackage{booktabs}
\usepackage{amsmath}
\usepackage{amsfonts}
\usepackage{graphicx}
\usepackage{xcolor}
\usepackage{amsfonts}
\usepackage{amsmath}
\usepackage{bbm}
\usepackage[ruled,vlined,linesnumbered]{algorithm2e}
\usepackage[normalem]{ulem}
\usepackage{soul}
\usepackage{xcolor}
\usepackage[inline]{enumitem}
\usepackage{siunitx}
\usepackage{float}

\usepackage[textsize=tiny, textwidth=1.3cm]{todonotes}

\usepackage{subcaption} 
\usepackage{caption}
\usepackage{pgfplots}
\pgfplotsset{compat=1.8}

\usepgfplotslibrary{statistics}
\makeatletter
\pgfplotsset{
    boxplot/hide outliers/.code={
        \def\pgfplotsplothandlerboxplot@outlier{}%
    }
}
\makeatother

\SetKwRepeat{Do}{do}{while}
\usepackage{color, colortbl}
\usepackage{bbold}

\usepackage{flushend}
\setcopyright{ifaamas}  
\copyrightyear{2020} 
\acmYear{2020} 
\acmDOI{} 
\acmPrice{} 
\acmISBN{} 
\acmConference[AAMAS'20]{Proc.\@ of the 19th International Conference on Autonomous Agents and Multiagent Systems (AAMAS 2020)}{May 9--13, 2020}{Auckland, New Zealand}{B.~An, N.~Yorke-Smith, A.~El~Fallah~Seghrouchni, G.~Sukthankar (eds.)}  

\begin{document}

\title[On Algorithmic Decision Procedures in Emergency Response Systems]{On Algorithmic Decision Procedures in Emergency Response Systems in Smart and Connected Communities}  



%
\author{Geoffrey Pettet}
\affiliation{%
  \institution{Vanderbilt University}
  \city{Nashville}
  \state{TN}
  \postcode{37212}
}
\email{geoffrey.a.pettet@vanderbilt.edu}

\author{Ayan Mukhopadhyay}
\affiliation{%
  \institution{Stanford University}
  \city{Palo Alto}
  \state{CA}
  \postcode{94305}
}
\email{ayanmukh@stanford.edu}

\author{Mykel Kochenderfer}
\affiliation{%
  \institution{Stanford University}
  \city{Palo Alto}
  \state{CA}
  \postcode{94305}
}
\email{mykel@stanford.edu}

\author{Yevgeniy Vorobeychik}
\affiliation{%
  \institution{Washington University}
  \city{St Louis}
  \state{MO}
  \postcode{37212}
}
\email{yvorobeychik@wustl.edu}

\author{Abhishek Dubey}
\affiliation{%
  \institution{Vanderbilt University}
  \city{Nashville}
  \state{TN}
  \postcode{37212}
}
\email{abhishek.dubey@vanderbilt.edu}

\begin{abstract}  
Emergency Response Management (ERM) is a critical problem faced by communities across the globe. Despite this, it is common for ERM systems to follow myopic decision policies in the real world. Principled approaches to aid ERM decision-making under uncertainty have been explored but have failed to be accepted into real systems. We identify a key issue impeding their adoption --- algorithmic approaches to emergency response focus on reactive, post-incident dispatching actions, i.e. optimally dispatching a responder \textit{after} incidents occur. However, the critical nature of emergency response dictates that when an incident occurs, first responders always dispatch the closest available responder to the incident. We argue that the crucial period of planning for ERM systems is not post-incident, but between incidents. This is not a trivial planning problem --- a major challenge with dynamically balancing the spatial distribution of responders is the complexity of the problem. An orthogonal problem in ERM systems is planning under limited communication, which is particularly important in disaster scenarios that affect communication networks. We address both problems by proposing two partially decentralized multi-agent planning algorithms that utilize heuristics and exploit the structure of the dispatch problem.  We evaluate our proposed approach using real-world data, and find that in several contexts, dynamic re-balancing the spatial distribution of emergency responders reduces both the average response time as well as its variance. 

\end{abstract}


\pagenumbering{arabic}
\maketitle

\section{Introduction}
\label{sec:intro}

Emergency response management (ERM) is a critical problem faced by communities across the globe. First responders must attend to many incidents dispersed across space and time using limited resources. ERM can be decomposed into the following sub-problems --- forecasting, planning, and dispatching. Although these have been examined independently, planning and dispatch decisions are dependent on accurate incident forecasting. Therefore, it is imperative that principled approaches are designed to tackle all three sub-problems. However, it is fairly common for ERM systems to follow myopic and straight-forward decision policies. For decades, the most common dispatching approach was to send the closest available responder to the incident (in time or space), after which the responder would return to its base or be reassigned. Such methods do not necessarily minimize expected response times \cite{mukhopadhyayAAMAS17}. As cities grow, population density, traffic dynamics and the sheer frequency of incidents make such methods stale and inaccurate. We systematically investigate the nuances of algorithmic approaches to ERM and describe how principled decision-making can aid emergency response. 




Naturally, algorithmic approaches to emergency response typically combine a data-driven forecasting model to predict incidents with a decision-making process that provides dispatch recommendations. Canonical approaches towards modeling the decision process involve using a Continuous-Time Markov Decision Process (CT-MDP)\cite{keneally2016markov} or a Semi-Markovian Process (SMDP)\cite{mukhopadhyayAAMAS18}, which are solved through dynamic programming. While the SMDP model provides a more accurate representation of ERM dynamics, it does not scale well for dynamic urban environments\cite{mukhopadhyay2019online}. The trade-off between optimality and computational time has also been investigated by the use of Monte-Carlo based methods\cite{mukhopadhyay2019online}. 

Despite such algorithmic progress and attention in recent years from the AI community\cite{AIEmergency,mukhopadhyayAAMAS18,pettet2017incident,mukhopadhyay2019online,yue2012efficient,purohit18}, there are still issues that impede the adoption of principled algorithmic approaches. We argue that a major problem lies in the very focus of most algorithmic approaches. Most ERM systems seek to perform decision-making \textit{after} incidents occur. While such approaches guarantee optimality in the long run (with respect to response times), they de-prioritize response to some incidents. Our conversations with first-responders\cite{fireDepartmentCommunication} revealed two crucial insights about this problem: 1) it is almost impossible to gauge the severity of an incident from a call for assistance and de-prioritize immediate response in anticipation of higher future rewards, and 2) Computer-Aided Dispatch systems (CAD)\cite{wiki:cad} typically enable a human agent to dispatch a responder in the span of 5-10 seconds.
These insights explain why the closest responder is usually dispatched to an incident; it is too risky to de-prioritize incidents of unknown severity. 

We raise an important conceptual question about algorithmic approaches to emergency response - is it feasible to optimize over dispatch decisions once an incident has happened? In this paper we argue that the crucial, practical period of principled decision-making is \textit{between} incidents. This avoids the potential consequences of explicitly choosing to de-prioritize response to an incident to achieve future gain, but accommodates the scope of principled decision-making. Most ERM systems do not exploit the scope of dynamically \textit{rebalancing} the spatial distribution of responders according to the need of the hour. This problem is challenging since optimizing responder distribution and response as a multi-objective optimization problem is usually computationally infeasible. Indeed, even Monte-Carlo based methods have previously been used with a restricted action space (only responding to incidents) to achieve acceptable computational latency\cite{mukhopadhyay2019online}. We address this challenge by proposing two efficient algorithmic approaches to optimize over the spatial distribution of responders dynamically.

The second set of problems that impedes the adoption of algorithmic decision-making in ERM is related to resilience and efficiency. Data processing and decision-making for algorithmic dispatching usually occur in a centralized manner (typically at a central data processing center), which is then communicated to responders. 
ERM, however, clearly evolves in a multi-agent setting, in which the agents have the capacity to perform independent computation (most modern ambulances are equipped with laptops). In an extremely time-critical setting, especially during communication breakdowns often caused by disasters, it is crucial that such computing abilities are used, and distributed and parallelized algorithmic frameworks are designed. Also, centralized decision-making systems treat all agents as part of a monolithic object or state. This is redundant, as agents often operate independently (for example, an ambulance in one part of the city is usually not affected by an incident in a completely different or distant part). In this paper, we argue that decentralized planning could identify and utilize structure in the problem and save vital computational time.

 {\bf Contributions:} We focus on two problems in this paper
 \begin{enumerate*}
 \item designing an approach that can accommodate rebalancing of resources to ensure efficient response, and 
 \item designing the ability for an emergency response system to be equipped to deal with scenarios that require decentralized planning with very limited communication.
 \end{enumerate*} 
  To this end, we start by modeling the problem of optimal response as a Multi-Agent Semi-Markov Decision Process (M-SMDP)\cite{boutilier1996planning,ghavamzadeh2006learning}. Then, we describe a novel algorithmic approach based on Multi-Agent Monte-Carlo Tree Search (M-MCTS)\cite{claes2017decentralised} that facilitates parallelized planning to dynamically rebalance the spatial distribution of responders. Our approach utilizes the computation capacity of each individual agent to create a \textit{partially} decentralized approach to planning. Finally, we evaluate our framework using real-world data from Nashville, TN. We find that these approaches maintain system fairness while decreasing the average and variance of incident response times when compared to the standard procedure.
  
  \textbf{Outline:}  Through the rest of the paper, we describe the overall problem of emergency response and explain the algorithmic framework. We begin by providing a brief background regarding how ERM pipeline can be modeled technically, and how theoretical approaches to solution work in such situations. Then, we describe our algorithmic framework in detail, and finally, evaluate our framework using incident and response data from Nashville, TN. Table \ref{tab:lookup-table} can be used as a reference for the symbols we use.
\definecolor{Gray}{gray}{0.9}
\begin{table}[]
\caption{Notation lookup table}
\vspace{-0.1in}
\footnotesize
\captionsetup{font=small}
\resizebox{.95\columnwidth}{!}{%
\begin{tabular}{|l|l|}
\hline
\rowcolor{Gray}
{\ul \textbf{Symbol}}     & {\ul \textbf{Definition}}                                                                       \\ \hline
$\Lambda$                 & Set of agents                                                                                   \\ \hline\rowcolor{Gray}
$D$                       & Set of depots                                                                                   \\ \hline
$\mathcal{C}(d)$                    & Capacity of depot $d$                                                                           \\ \hline\rowcolor{Gray}
$G$                       & Set of cells                                                                                    \\ \hline
$S$                       & State space                                                                                     \\ \hline\rowcolor{Gray}
$A$                       & Action space                                                                                    \\ \hline
$P$                       & State transition function                                                                       \\ \hline\rowcolor{Gray}
$T$                       & Temporal transition distribution                                                                \\ \hline
$\alpha$                  & Discount factor                                                                                 \\ \hline\rowcolor{Gray}
$\rho(s, a)$                    & Reward function given action $a$ taken in state $s$                                                                                \\ \hline
$\mathcal{A}$             & Joint agent action space                                                                        \\ \hline\rowcolor{Gray}
$\mathcal{T}$             & Termination scheme                                                                              \\ \hline
$s^t$                     & Particular state at time $t$                                                                    \\ \hline\rowcolor{Gray}
$I^t$                     & Set of cell indices waiting to be serviced                                                     \\ \hline
$R^t$                     & Set of agent states at time $t$                                                                 \\ \hline\rowcolor{Gray}
$p^t_j$                   & Position of agent $j$                                                                           \\ \hline
$g^t_j$ & Destination of agent $j$                                                                        \\ \hline\rowcolor{Gray}
$u^t_j$ & Current status of agent $j$                                                                     \\ \hline
$s_i, s_j$                & Individual states                                                                               \\ \hline\rowcolor{Gray}
$t_{ij}$                    & Transition time between states $s_i, s_j$                                                       \\ \hline
$t_a$                     & Time between incidents                                                                          \\ \hline\rowcolor{Gray}
$t_s$                     & Time to service an incident                                                                     \\ \hline
$t_b$                     & Time to a balance step                                                                          \\ \hline\rowcolor{Gray}
$r$                       & Reward                                                                                          \\ \hline
$t_r$                     & \begin{tabular}[c]{@{}l@{}}Incident response time (the time between incident \\ awareness and first agent's arrival on scene)\end{tabular} \\ \hline\rowcolor{Gray}
$\sigma$                  & Action recommendation set                                                                        \\ \hline
$\mu$                     & Mean agent service time                                                                         \\ \hline\rowcolor{Gray}
$c_d$                     & Number of agents at depot $d$                                                                   \\ \hline
$\upsilon_g$              & Incident rate at cell $g$                                                                       \\ \hline\rowcolor{Gray}
$\upsilon_g^d$            & The fraction of cell $g$'s incident rate shared by depot $d$                                    \\ \hline
$\Upsilon$                & Set of occupied depots and their split incident rates                                           \\ \hline\rowcolor{Gray}
$\pi_\Upsilon$            & Utility of $\Upsilon$                                                                           \\ \hline
$t_h$                     & Time since beginning of planning horizon                                                        \\ \hline\rowcolor{Gray}
$t_r(s,a)$                   & Response time to an incident given action $a$ in state $s$                                      \\ \hline
$\phi_k(s,a)$                & Distance traveled by agent $k$ while balancing                                                  \\ \hline\rowcolor{Gray}
$\psi$                    & \begin{tabular}[c]{@{}l@{}} Exogenous parameter balancing response time \\ and distance traveled 
\end{tabular}                               \\ \hline
RoI& Radius of Influence of a cell (used in queue based rebalancing policy). 
\\ \hline
\end{tabular}%
}
\vspace{-0.2in}
\label{tab:lookup-table}
\end{table}

\section{System Model}\label{sec:model}
Our goal is to develop an approach for emergency responder placement and incident response in a dynamic, continuous-time and stochastic environment. We begin with several assumptions on the problem structure and information provided \textit{a-priori}. First, we assume that we are given a spatial map
broken up into a finite collection of equally-sized cells $G$, and that we
are given an exogenous spatial-temporal model of incident arrival in continuous time over this collection of cells (we describe one such model later). Second, we assume that for each spatial cell, the temporal distribution of incidents is homogeneous. 
Our third assumption is that emergency responders are allowed to be housed in a set of fixed and exogenously specified collection of depots $D$. Depots are essentially a subset of cells that responders can wait in, and are analogous to fire-stations in the real-world. Each depot $d \in D$ has a fixed capacity $\mathcal{C}(d)$ of responders it can accommodate at a time. We assume that when an incident happens, a free responder (if available) is dispatched to the site of the incident. Once dispatched, the time to service consists of two parts: 1) time taken to travel to
the scene of the incident, and 2) time taken to attend to the incident. If no free responders are available, then the incident enters a waiting
queue. 

\subsection{Incident Arrival}\label{sec:survival}

An important component of a decision-theoretic framework to aid emergency response is the understanding of \textit{when} and \textit{where} incidents occur. 
While our algorithmic framework can work with any forecasting model, we briefly describe the one that we choose to use: a continuous-time forecasting model based on survival analysis. It has recently shown state-of-the-art performance in prediction performance for a variety of spatial-temporal incidents (crimes, traffic accidents etc.)\cite{mukhopadhyayAAMAS17,mukhopadhyayGameSec16,pettet2017incident}.
Formally, the model represents a probability distribution over inter-arrival times between incidents, conditional on a set of features, and can be represented as 

\[
    f_t(T=t|\gamma(w))
\]

where $f_t$ is a probability distribution for a continuous random 
variable $T$ representing the inter-arrival time, which typically depends on
covariates $w$ via the function $\gamma$. The model parameters can be estimated by the principled procedure of Maximum Likelihood Estimation (MLE) \cite{guo2010survival}.


\subsection{Decision-Making Process}

The evolution of incident arrival and emergency response occur in continuous-time, and can be cohesively represented as a Semi-Markov Decision Process (SMDP) \cite{mukhopadhyayAAMAS18}.
An SMDP system can be described by the tuple
 $(S,A,P,T,\rho(i,a),\alpha)$ where $S$ is a finite state space, $A$ is the set of actions, $P$ is the state transition function with $p_{ij}(a)$ being the probability with which the process transitions from state $i$ to state $j$ when action $a$ is taken, $T$ denotes the temporal transition with $t(i,j,a)$ representing a distribution over the time spent during the transition from state $i$ to state $j$ under action $a$, $\rho$ represents the reward function, and $\alpha$ is the discount factor.

To adapt this formulation to a multiagent setting, 
 we model the evolution of incidents and responders together in a Multi-Agent SMDP (MSMDP)\cite{rohanimanesh2003learning}, which can be represented as the tuple $(\Lambda,S,\mathcal{A},P,T,\rho(i,a),\alpha,\mathcal{T})$, where $\Lambda$ is a finite collection of agents and $\lambda_j \in \Lambda$ denotes the $j^{\text{th}}$ agent. The action space of the $j^{\text{th}}$ agent is represented by $A_j$, and $\mathcal{A} = \prod_{i=1}^{m} A_j$ represents the joint action space. We assume that the agents are cooperative and work to maximize the overall utility of the system. The components  $S$, $\rho$ and $P$ are defined as in a standard SMDP. $\mathcal{T}$ represents a termination scheme; note that since agents each take different actions that could take different times to complete, they may not all terminate at the same time. An overview of such schemes can be found in prior literature \cite{rohanimanesh2003learning}. We focus on asynchronous termination, where actions for a particular agent are chosen as and when the agent completes it's last assigned action. Next, we define the important components of the decision process in detail.
 
 \textbf{States:}
A state at time $t$ is represented by $s^t$ which consists of a tuple $(I^t, R^t)$, where $I^t$
is a collection of cell indices that are waiting
to be serviced, ordered according to the relative times of incident occurrence. $R^t$ corresponds to information about the set of agents at time $t$ with $|R^t| = |\Lambda|$.
Each entry $r^t_j \in R^t$ is a set $\{p^t_j,g^t_j,u^t_j\}$, where $p^t_j$ is the position of responder $\lambda_j$, $g^t_j$ is the destination cell that it is traveling to (which can be its current position), and $u^t_j$ is used to encode its current status (busy or available), all observed at the state of our world at time $t$.
For the sake of convenience, we abuse notation slightly and refer to an arbitrary state simply by $s$ and use the notation $s_i$ and $s_j$ to refer to multiple states. We point out that our model revolves around states with specific events that provide the scope of decision-making. Specifically, decisions need to be taken when incidents occur, when responders finish servicing and while rebalancing the distribution of responders. We also make the assumption that no two events can occur simultaneously in our world. In case such a scenario arises, since the world evolves in
continuous time, we can add an arbitrarily small time interval to segregate the two events and create two separate states.

\textbf{Actions:}
Actions in our world correspond to directing the responders to a valid cell to either respond to an incident or wait.  Valid locations include cells with pending incidents or any depot that has capacity to accommodate additional responders. For a specific agent $\lambda_i$, valid actions for a specific state $s_i$ are denoted by $A^i(s_i)$ (some actions are naturally invalid, for example, if an agent is at cell $k$ in the current state, any action not originating from cell $k$ is unavailable to the agent). Actions can be broadly divided into two categories - \textit{responding} and \textit{rebalancing}. Responding actions refer to an agent actually going to the scene of an incident to service it. But agents could also be directed to wait at certain depots based on the likelihood of future incidents in the proximity of the said depot. We refer to such actions as rebalancing. Finally, we reiterate that the joint valid action space of all the agents and a particular instantiation of it are defined by $\mathcal{A}$ and $a$ respectively, and that of a specific agent $\lambda_j$ by $A_j$ and $a_j$.

\textbf{Transitions:}
Having described the evolution of our world, we now look at
both the transition time between states, as well as the probability of observing a state, given the last state and action taken. We define
the former first, denoting the time between two states $s_i$
and $s_j$ by
the random variable $t_{ij}$. There are four random variables of interest in this context. We denote the time between incidents by the random variable $t_a$, the time to service an incident by $t_s$, the time taken for a balancing step as $t_b$ and the time taken for a responder to reach the scene of an incident by $t_r$. We overload these notations for convenience later.
Specifically, we model $t_a$ using a survival model described in section \ref{sec:survival}. We model the service times ($t_s$) by learning an exponential distribution from service times using historical emergency response data,
and we model rebalancing time ($t_b$) simply by the time taken by an agent to move to the directed cell.

We refrain from focusing on the transition function $P$, as our algorithmic framework only needs a generative model of the world and not explicit estimates of state transition probabilities.


\textbf{Rewards:}
Rewards in SMDP usually have two components: a lump sum instantaneous reward for taking actions, and a continuous time reward
as the process evolves. Our system only involves the former, which
we denote by $\rho(s, a)$, for taking action $a$ in state $s$. We define the exact reward function in section \ref{sec:other_agents_models}. 





\subsection{Problem Definition} \label{sec:problem_def}
 Given state $s$ and an agent set $\Lambda$, the problem is to determine  an action recommendation set $\sigma = \{a_1,...,a_m\}, \,\,s.t.\,\, a_i \in A^i(s)$, that maximizes the expected reward. The $i$th entry in $\sigma$ contains a \textit{valid} action for the $i$th agent. 
 
 Sovling this problem directly is hard due to its intractable state space. Further, the state transition functions are unknown and difficult to model in closed form, which is typical of urban scenarios where incidents and responders are modeled cohesively \cite{mukhopadhyayAAMAS18}. Finally, we have to consider the following practical constraints and limitations.
\begin{itemize}[noitemsep,leftmargin=5.5mm]
    \item Temporal constraints --- emergency response systems can afford minimum latency (~5-10 seconds in practice).
    \item Capacity constraints --- each depot has a fixed agent capacity.
    \item Uniform severity constraint --- all incidents must be responded to `promptly', without making a judgement about its severity based on a report or a call.
    \item Wear and Tear --- The overall distance agents travel should be controlled to limit vehicle wear and tear.
    \item Limited Communication - ERM systems must be equipped to deal with disaster situations, where communication is limited. 
\end{itemize}
The temporal and uniform severity constraints make it difficult to justify implementing dispatch policies other than greedy; in order to improve upon greedy dispatch, some `good' myopic rewards must be sacrificed for an increase in expected future rewards. Since it is very hard to predict the severity of an incident pre-dispatch, the decision process cannot determine if this sacrifice is acceptable. Therefore, in this work we focus on \textit{inter-incident} planning while maintaining greedy dispatch decisions when an incident is reported. This approach gives the decision-maker more flexibility, as it can proactively position resources rather than reacting to incidents. Our problem then becomes how to distribute responders between incidents such that the greedy dispatching rewards are maximized. 



\section{Rebalancing Approach to ERM}

\subsection{Problem Complexity}

Dynamic rebalancing's flexibility comes with an increase in complexity. Consider an example city with $|\Lambda|$ responders (i.e. agents) and $|D|$ locations where responders could be stationed (called \textit{depots}) that each can hold one responder. When making a dispatch decision at the time of an incident, a decision maker has at most $|\Lambda|$ possible choices: which responder to dispatch. If instead it is re-assigning responders across depots, there are significantly more choices. For example, with $|\Lambda|=20$ and $|D|=30$, there are 20 dispatching choices per incident, but $P(|D|,|\Lambda|) = \frac{|D|!}{(|D|-|\Lambda|)!} = \frac{30!}{10!} = \num{7.31e+25}$ possible assignments. This will only increase if depots have higher responder capacities.  


Approaching the problem from this perspective requires solutions that can cope with this large complexity. One possible approach is to directly solve the SMDP model. Although the state transition probabilities are unknown, one can estimate the transition function by embedding learning into policy iteration\cite{mukhopadhyayAAMAS18}. This approach is unsuitable for rebalancing, as it is too slow even for the dispatch problem. A centralized MCTS approach suffers from the same shortcoming, barely satisfies the computational latency constraints in case of the dispatch problem\cite{mukh_pettet_2019_iccps}. Instead, we seek to exploit meaningful heuristics to propose computationally feasible rebalancing strategies. We begin by presenting our first approach, which focuses on using historical frequencies of incident occurrence across cells to assign responders. 








\subsection{Multi-Server Queue Based Rebalancing}\label{sec:rate_rebal_description}
One way to address the complexity of rebalancing is by considering an informed heuristic. A natural heuristic for ERM rebalancing is \textit{incident rate} --- each depot can be assigned responders based on the total rate of incidents it serves. Ultimately, our goal is to find a rebalancing strategy that minimizes expected response times. As a result, we first estimate the response time given a specific assignment of responders. Such a scenario can be modeled as a multi-server M/M/c queue \cite{gautam2012analysis}. For a given cell and depot, the response time for an M/M/c queue can be represented as  

\begin{equation}
\begin{aligned}
&\text{responseTime}(c_d, \upsilon, \mu) = \frac{\omega(c_d, \upsilon/\mu)}{c_d\mu - \upsilon} + \frac{1}{\mu} \label{eq:queue_resp_time}\\
&\text{where}\,\,\,
\omega(c_d, \upsilon / \mu) = \frac{1}{1 + (1-\frac{\upsilon}{c_d\mu})(\frac{c_d!}{(c_dq)^c})\sum_{k=0}^{c_d-1}\frac{(c_dq)^k}{k!}}
\end{aligned}
\end{equation}
where $\mu = \mathbb{E}(t_s)$ is the mean service time of responders, $c_d$ is the number of responders stationed at the depot, $\upsilon$ denotes the rate of incident occurrence at the concerned cell, and $q = \frac{\upsilon}{c \mu} $ is server utilization. The standard M/M/c model above needs slight adjustment to account for the fact that incidents at a cell $g$ can potentially be serviced by any depot, which are located at different distances from $g$. Therefore, we consider a multi-class queue formulation in which a cell's incident rate is split among each depot. Since depots closer to a cell $g$ are more likely to service its incidents, we split $g$'s incident arrival rate such that the fraction of rate incurred by a depot is inversely proportional to the distance to $g$.


The following system of linear equations can be used to split the arrival rate of a cell $g$ among depots $D$.

\begin{subequations}
\begin{align}
&\sum_{d \in D}\upsilon_g^d =  \upsilon_g \label{eq:weightedLambdas}\\ 
&\text{dist}(\widetilde{d}, g)\upsilon_g^{\widetilde{d}} = \text{dist}(d_i, g)\upsilon_g^{d_i} \,\,\,\,\forall d_i \in D \backslash \widetilde{d} \label{eq:distLambda}
\end{align}
\end{subequations}

where the variable $\upsilon_g^d$ is the fraction of arrival rate of cell $g$ that is shared by depot $d$, $\text{dist}(d,g)$ denotes the distance between depot $d$ and cell $g$, and $\widetilde{d}$ is the depot closest to $g$. Equation \ref{eq:weightedLambdas} ensures that the split rates for each cell $g \in G$ sum to its actual arrival rate $\upsilon_g$, and equation \ref{eq:distLambda} ensures that the weighted $\upsilon$'s are inversely proportional to the relative distances between the depots and the cell. For convenience, we refer to the entire set of split rates by $\Upsilon$.



The split rates $\Upsilon$ provide a foundation for a responder rebalancing approach, given a few considerations. First, we might not have enough responders to meet the total demand based on $\Upsilon$. Secondly, the problem of evaluating response times in the context of emergency response is different than the standard M/M/c queue formulation, since travel times are not memoryless, and must be modeled explicitly.
To address these issues, we design a scoring mechanism for evaluating a specific allocation of responders to depots for a given $\Upsilon$. We denote this score by $\pi_\Upsilon$.
Using $\Upsilon$, a responder allocation can be scored by summing each depot $d$'s expected response time based on the queuing model (calculated using equation \ref{eq:queue_resp_time}) and the overall time taken by responders to complete the rebalancing: 

\begin{equation}
    \pi_{\Upsilon} = \sum_{d \in D} \sum_{g \in G} \mathbb{1}(d,\Lambda)\{\textit{responseTime}(c_d, \upsilon_g^d, \mu) + \textit{travelTime}(d, g)\}
\end{equation}
where $\mathbb{1}(d,\Lambda)$ is an indicator function which set to $1$ only if depot $d$ has at least one responder, and the functions $responseTime$ and $travelTime$ are used to denote the expected response time of a depot and travel times needed by agents to respond to incidents. The goal of an assignment method is then to find a responder allocation that minimizes this heuristic score. To minimize the total score
we employ an iterative greedy approach, shown in algorithm \ref{alg:action_select}. Once the best depots are found, responders are assigned to them based on their current distance from the depots. 







\begin{algorithm}[t]
\footnotesize
\caption{Iterative Greedy Action Selection}
\label{alg:action_select}
\textbf{INPUT}:  number of agents $|\Lambda|$, depots $D$, depot capacities $C$, grid rates $\upsilon_g \forall g \in G$\;

final\_depot\_occupancy := Hash \{$d: 0$\} $\forall d \in D$  \;

\Do{$sum($final\_depot\_occupancy$) < |\Lambda|$}
{

candidate\_depots := Set $\emptyset$\; candidate\_scores := Hash $\emptyset$\;

\For{$d \in D$}
{

\If{final\_depot\_occupancy[$d$] $< C(d)$}{

temp\_occ := final\_depot\_occupancy\; temp\_occ[$d$] $+= 1$\;
find $\Upsilon_d$  by solving system of linear equations \{\ref{eq:weightedLambdas}, \ref{eq:distLambda}\} given temp\_occ\;
$\pi_{\Upsilon_d} := \sum_{d \in D} \sum_{g \in G} \mathbb{1}(d,\Lambda)\{\textit{responseTime}($temp\_occ[$d$]$, \upsilon_g^d, \mu) + \textit{travelTime}(d, g)\} $\;

candidate\_depots := candidate\_depots $\cup d$\; candidate\_scores := candidate\_scores $\cup \{d: \pi_{\Upsilon_d}\}$ 

}}
best\_depot := argmin $\pi_{\Upsilon_{d}} \,\,\,\, \forall d \in$ candidate\_depots\;
final\_depot\_occupancy[best\_depot] $+= 1$\;

}
return chosenDepots\;
\end{algorithm}


The approach dramatically decreases the computational complexity of rebalancing compared to a brute force search. The complexity for solving the system of linear equations \{\ref{eq:weightedLambdas}, \ref{eq:distLambda}\} is $\mathcal{O}(|\Lambda|^3)$, as there are at most $|\Lambda|$ depots that could have a resource allocated. The rates are split for each cell $g \in G$ and new depot under consideration $d$ during each iteration of the greedy search in algorithm \ref{alg:action_select}, which is repeated $|\Lambda|$ times to place each responder. This gives the overall algorithm a complexity of $\mathcal{O}(|G||D||\Lambda|^5)$. Taking the same example given above with $|\Lambda|=20$ and $|D|=30$ and assuming $|G| = 900$ (based on our geographic area of interest and patrol areas chosen by local emergency responders), the complexity is $\num{1e+15}$ times less than a brute force search. 

While this approach is not inherently decentralized, each agent can perform these computations and take actions themselves, requiring minimal coordination. While straightforward and tractable, there are a few potential downsides to this approach. First, this policy does not take into account the internal state of the system. For example, a responder might be on its way to respond to an incident, thereby rendering it unavailable for rebalancing. Secondly, it assumes that historical rates of incident arrival can be used to optimize responder placement for the future, thereby not considering how future states of the system affect a particular rebalancing configuration. To address these issues, we propose a decentralized Monte-Carlo Tree Search algorithm.

\subsection{Decentralized MCTS Approach}
\label{sec:other_agents_models} 





Monte-Carlo Tree Search (MCTS) is a simulation-based search algorithm that has been widely used in game playing scenarios.

MCTS based algorithms evaluate actions by sampling from a large number of possible scenarios. The evaluations are stored in a search tree, which is used to explore promising actions. Typically, exploration policy is dictated by a principled approach like UCT\cite{kocsis2006bandit}. A standard MCTS-based approach is not suitable for our problem due to the sheer size of the state-space in consideration coupled with the low latency that ERM systems can afford. Instead, we focus on a decentralized multi-agent MCTS (MMCTS) approach explored by Claes et. al \cite{claes2017decentralised} for multi-robot task allocation during warehouse commissioning. 
In MMCTS individual agents build separate trees focused on their own actions, rather than having one monolithic, centralized tree. This dramatically reduces the search space: in our case, at each evaluation step of a Monte-Carlo based approach, using a decentralized multi-agent search reduces the total number of choices from the number of permutations $P(|D|,|\Lambda|) = \frac{|D|!}{(|D|-|\Lambda|)!}$ to only the number of depots $|D|$. 

To realize MMCTS for an ERM domain, some extensions need to be made to standard UCT \cite{furnkranz2006machine}. While an agent is building its own tree, it must model other agents' behavior. Since this estimation is required at every step of every simulation by each agent, finding a model that strikes a balance between computation time and accuracy of predicted actions is vital. 

There are also global constraints on the system which mandate agents maintain a minimal degree of coordination. For example, the number of resources assigned to a depot cannot be higher than its capacity. We take this into account by adding a filtering step to the decision process. Similar to Map-Reduce \cite{Dean:2008:MSD:1327452.1327492}, each agent sends their evaluated actions to a central planner which makes the final decisions while satisfying global system constraints. 

Next, we describe the architecture of our decentralized MMCTS based algorithm.

\begin{itemize}[noitemsep,leftmargin=*]
    \item  \textbf{Reward Structure:} At the core of an MCTS approach is an evaluation function that measures the reward of taking an action in a given state. For a state $s$ in the tree of agent $\lambda_j$, we design the reward $\rho$ of taking an action $a$ in $s$ as 

\begin{subequations}\label{eq:running action_util}
    

    \begin{align}
        \rho(s,a) = 
        \begin{cases}
        \rho_{s-1} - \alpha^{t_h}(t_{r}(s,a)) ,& \text{if responding to an incident} \\
        \rho_{s-1} - \alpha^{t_h}\psi\frac{ \sum_{\lambda_k \in \Lambda} (\phi_k(s,a))}{|\Lambda|}, & \text{if balancing at $s$}
        \end{cases}
    \end{align}
\end{subequations}

where $\rho_{s-1}$ refers to the total accumulated reward at the parent of state $s$ in the tree, $\alpha$ is the discount factor for future rewards, and $t_h$ the time since the beginning of the planning horizon $t_{0}$. The evaluation function is split into cases reflecting the separate \textit{incident dispatch} and \textit{balancing} steps in our solution approach. In a dispatch step, the reward is updated with the discounted response time to the incident $t_{r}(s,a)$. 
In a balancing step, we update the reward by the average distance traveled by the agents (we denote the distance traveled by agent $\lambda_k$ while balancing due to action $a$ in $s$ by $\phi_k(s,a)$). $\psi$ is an exogenous parameter that balances the trade-off between response time and distance traveled for balancing, and is set by the user depending on their priorities. Distance is not included during dispatch actions, as we always send the closest agent. 


\item \textbf{Evaluating other agents' actions:} 
Agents must have an accurate yet computationally cheap model of other agents' behavior; we explore two such possible policies --- (1) a naive policy that other agents will not rebalance, remaining at their current depot (referred to as \textit{Static Agent Policy}), and (2) an informed policy, which is in the form of the \textit{Queue Rebalancing Policy} described in the section \ref{sec:rate_rebal_description}. These are used to select actions for the other agents $\Lambda \backslash \{\lambda_i\}$ when building agent $\lambda_i$'s search tree, and are represented by the ActionSelection({available agents}, state) function in line $5$ of algorithm \ref{alg:expand}.

\item \textbf{Rollout:} When working outside the MMCTS tree, i.e. rolling out a state, a fast heuristic is used to estimate the score of a given action. We use greedy dispatch without balancing as our heuristic. 

\item\textbf{Action Filtering:}
The dispatching domain has several global constraints to adhere to, including ensuring that an incident is serviced if agents are available and that depots are not filled over capacity. To meet these constraints, we propose a filtering step be added to the MMCTS workflow, similar to Map-Reduce. Once each individual agent has scored and ranked each possible action, these are sent to a centralized filter that chooses the final actions for each agent to maximize utility without breaking any constraints. 
\end{itemize}

Another way global constraints affect the workflow is that the set of valid actions for an agent when they build their search tree may not be the same as the valid actions when it comes time for them to make a decision. For example, consider two agents $\lambda_1$ and $\lambda_2$; if agent $\lambda_1$ moves to a station and fills it to capacity, then agent $\lambda_2$ cannot move to that station. To address this, we have agents evaluate every action they could possibly take when expanding nodes in the tree, even if those actions would cause an invalid state. As the filter assigns actions to other agents, some of these actions can become valid. 

\newlength{\textfloatsepsave} \setlength{\textfloatsepsave}{\textfloatsep} \setlength{\textfloatsep}{0pt} 
\begin{algorithm}[t]
\footnotesize
\caption{Decision Process}
\label{alg:decision_process}
\textbf{INPUT}: state $s$, time limit $t_{lim}$\;

$I$ := Sample Incidents(s)

$E$ := $I$ + rebalancing events

ranked action set $\widetilde{\mathcal{A}} := \emptyset$\;

\For{Agent $\lambda_j \in \Lambda$}
{
    $\widetilde{\mathcal{A}}[\lambda_j]$ := MMCTS(s, $\lambda_{j}$, $E$, $t_{lim}$)\;
}

recommended actions $\sigma$ := CentralizedActionFilter($s$, $\widetilde{\mathcal{A}}$)\;

apply $\sigma$ to $s$\;

Return s\; 

\end{algorithm}

\begin{algorithm}[t]
\footnotesize
\caption{MMCTS}
\label{alg:mmcts}
\textbf{INPUT}: state s, agent $\lambda_j$, sampled events $E$, time limit $t_{lim}$\;

create root of search tree at s\;
\Do{within time limit $t_{lim}$}
{
    select most promising node $n$ from tree using UCB1\;
    childNode := Expand($n$, $\lambda_j$, next event $e \in E$ after state($n$))\;
    $r_c$ := Rollout(childNode)\; 
    back-propagate(child, $r_c$)
}
return actions $\lambda_j$ could take ranked by average reward
\end{algorithm}

\begin{algorithm}[t]
\footnotesize
\caption{Expand}
\label{alg:expand}
\textbf{INPUT}: Search Tree Node $n$, agent $\lambda_j$, next important event $e$\;

\uIf{$e$ is balancing step}{
    select un-explored action $a \in A_j$ \; 
    $\lambda_j$ takes action $a$\;
    actions available to other agents are updated
    ActionSelection($\Lambda \backslash \{\text{unavailable agents}\}$, state($n$))\; 
}\uElseIf{$e$ is an incident}{
    dispatch nearest agent to incident
}

create new child node $n_c$ from selected actions\; 
update the child's reward based on the response times (if any) and agent balancing movement

update $n_c$ to the time of the next event $e$, fast forwarding the state\; 

return $n_c$\;

\end{algorithm}









\begin{algorithm}[t]
\footnotesize
\caption{Centralized Action Filter}
\label{alg:filter}
\textbf{INPUT}:  state $s$, ranked actions $\widetilde{\mathcal{A}}$\;

$\Lambda_{avail}$ := agents($s$)
\Do{there are unassigned agents}
{
candidate\_actions := $\emptyset$\;
\For{Agent $\lambda_j \in \Lambda_{avail}$}{
 candidate\_actions[$\lambda_j$] := \textit{valid} action $a_j \in \widetilde{\mathcal{A}}[\lambda_j]$ with highest reward $\rho(s,a)$\;
}
find agent $\lambda_j$ with highest scored action $a_j \in $ candidate\_actions\;
$\lambda_j$ takes action $a_j$\;
update actions available to other agents accordingly\;
remove $\lambda_j$ from $\Lambda_{avail}$\;
}
\end{algorithm}

\setlength{\textfloatsep}{\textfloatsepsave}
\section{Integration Framework} \label{sec:experimental_setup}

\begin{figure}[t]
    \centering
    \includegraphics[width=0.8\columnwidth]{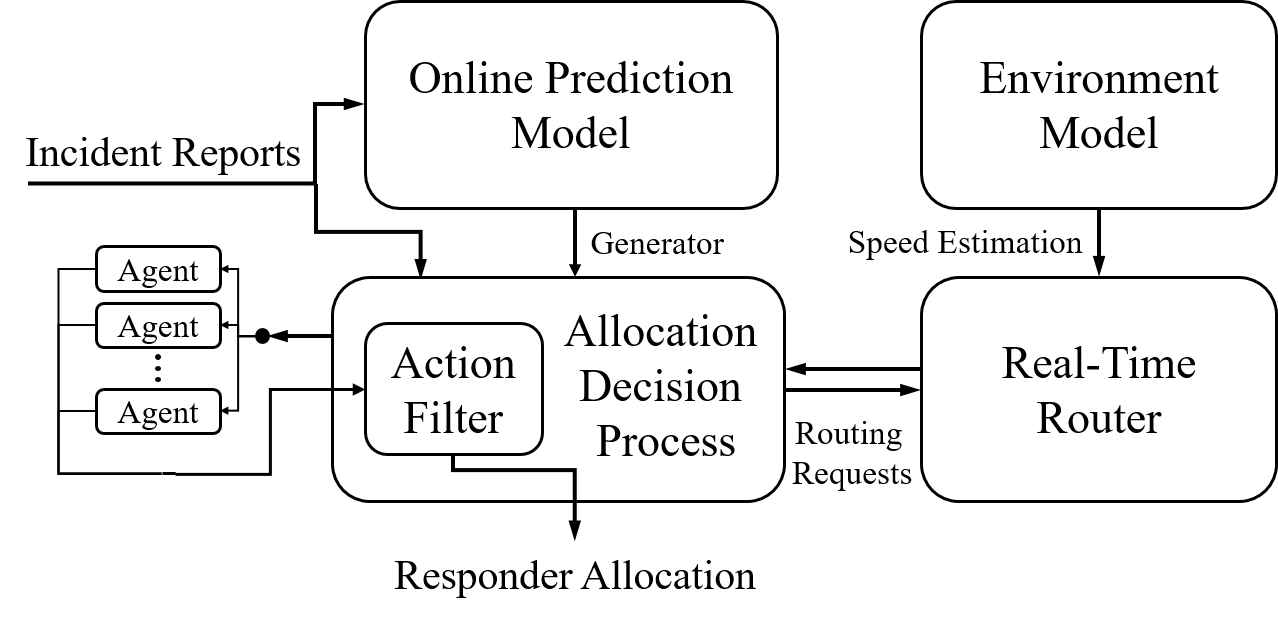}
    \vspace{-0.1in}
    \caption{Extended Decentralized ERM Framework Overview}
    \label{fig:new_framework}
    \vspace{-0.2in}
\end{figure}

To realize an online ERM decision support system requires a framework of interconnected processes. Our integration framework is built on our prior modular ERM pipeline work\cite{mukhopadhyay2019online}. 
It includes the following components: 


\begin{itemize}[noitemsep,leftmargin=5.5mm]
    \item A traffic routing model to support routing requests.
    \item A model of the environment and how it changes over time, which is used by the incident prediction model.
    \item A model of the spatio-temporal distribution of incidents.
    \item A decision process that makes dispatching recommendations based on the current state of the environment, responder locations, and future incident distributions. 
\end{itemize}
This framework is a natural choice as it decouples the decision process (our focus in this work) from other components. As it was designed for the centralized, post-incident dispatching approach, we make necessary changes to adapt it to our needs. The underlying discrete event simulation was generalized to accept events other than incident occurrence, such as periodic balancing events. The decision process was also extended to handle distributed, multi-agent approaches. An overview of the extended framework can be seen in figure \ref{fig:new_framework}.


In our experiments we use a Euclidean distance based router, and the incident prediction model outlined in section \ref{sec:model}. Due to the framework's modularity, these components can be replaced without affecting the decision process.

\begin{figure}[t]
\centering
\includegraphics[width=\columnwidth]{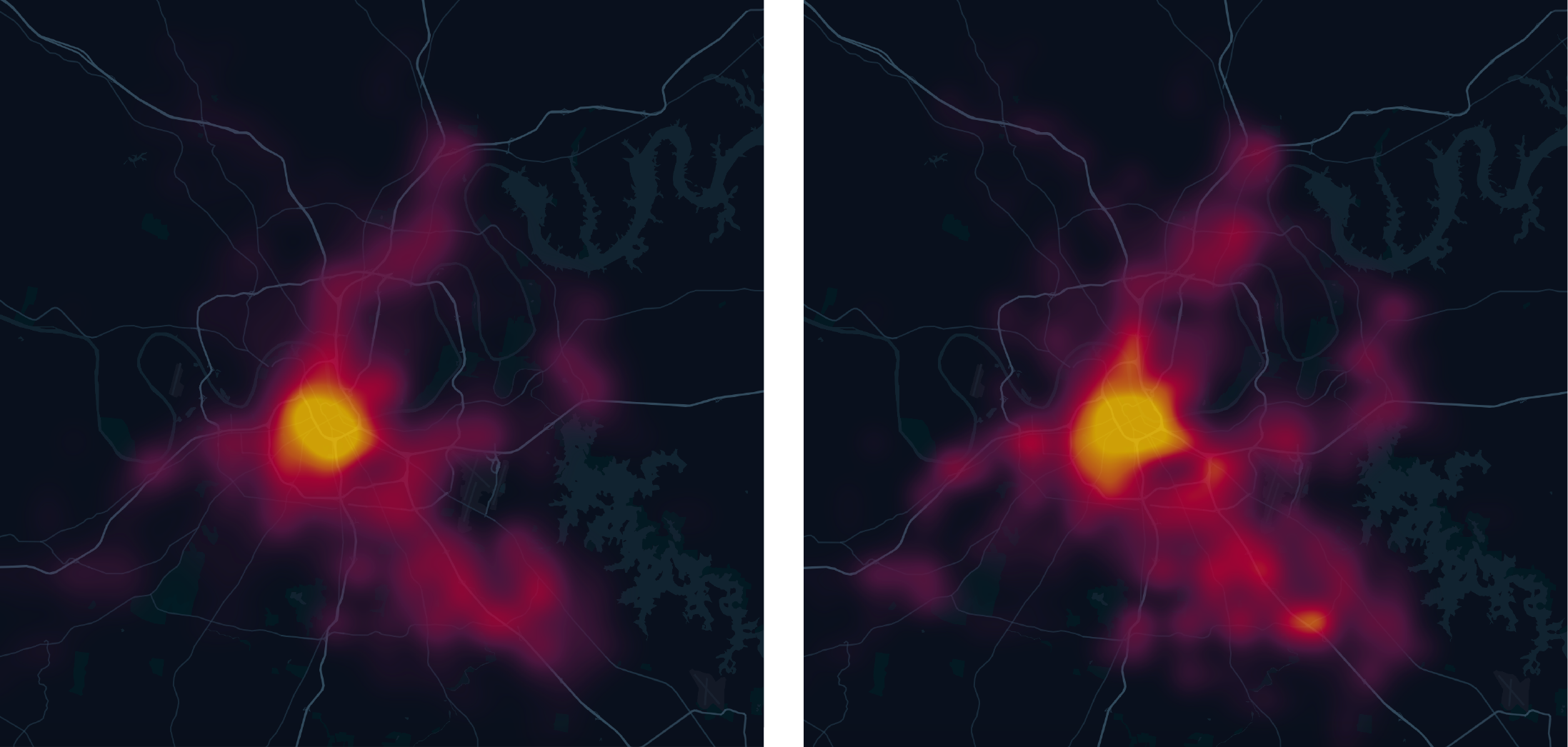}
\caption[caption for img]{Heatmaps comparing average incident rates for the forecasting model (left) with actual incidents in Nashville, TN (right)}.
\label{fig:comparisonforecast}
\vspace{-0.25in}
\end{figure}

\textbf{Incident Prediction Model:} \label{sec:incident_model}
While the broader approach of rebalancing the spatial distribution of responders is flexible enough to work with any modular incident forecasting model, we provide a brief evaluation of forecasting using survival analysis. To this end, we generate forecasts $4$ hours into the future at intervals of every half an hour for the entire test set, and then repeat the procedure 5 times to reduce variance and increase our confidence in the forecasts. Finally, we create a heatmap (average of all forecasted rates in the test set) to visualize the performance of the model in comparison to actual incidents (see figure \ref{fig:comparisonforecast}).
The forecasting models captures the high and low density areas fairly accurately, as well as the spatial spread of the incidents.


\subsection{Experimental Design}

We perform our evaluation on data from Nashville, TN, a major metropolitan area of USA, with a population of approximately 700,000. The depot locations are based on actual ambulance stations obtained from the city. Traffic accident data was obtained from the Tennessee Department of Transportation, and includes the location and time of each incident. The incident prediction model was trained on 35858 incidents occurring between 1-1-2018 and 1-1-2019, and we evaluated the decision processes on 2728 incidents occurring in the month of January, 2019.

\textbf{Experimental Configuration and Assumptions:}
We limit the capacity of each depot to 1 in our experiments. This is motivated by two factors --- first, it encourages responders to be geographically spread out to respond quickly to incidents occurring in any region of the city, and it models the usage of ad-hoc stations by responders, which are often temporary parking spots.
While the responder service times to incidents are assumed to be exponential in the real world, we set them to a constant for these experiments. This ensures that the experiments across different methods and parameters are directly comparable. If deployed, however, proper service time distributions should be learned and sampled from for each ERM system.  We set the total number of responders to 26, which is the actual number of responders in Nashville. We split the geographic area into 900, 1x1 mile square cells. This choice was a consequence of the fact that a similar granularity of discretization is followed by local authorities. To smooth out model noise, each agent evaluates 5 sampled incident chains from the generative model and averages the scores for each action across the playouts. The standard UCB1 \cite{auer2002finite} algorithm is used to select the most promising node during MCTS iterations. Finally, we augment the queue based rebalancing policy by adding a \textit{radius of influence (RoI)} for each cell. Only depots within a cell's RoI are considered when splitting its rate to encourage even agent distribution and reduce computation time.

\begin{figure*}[]
\begin{subfigure}{.163\linewidth}
  \centering
 \includegraphics[width=\linewidth]{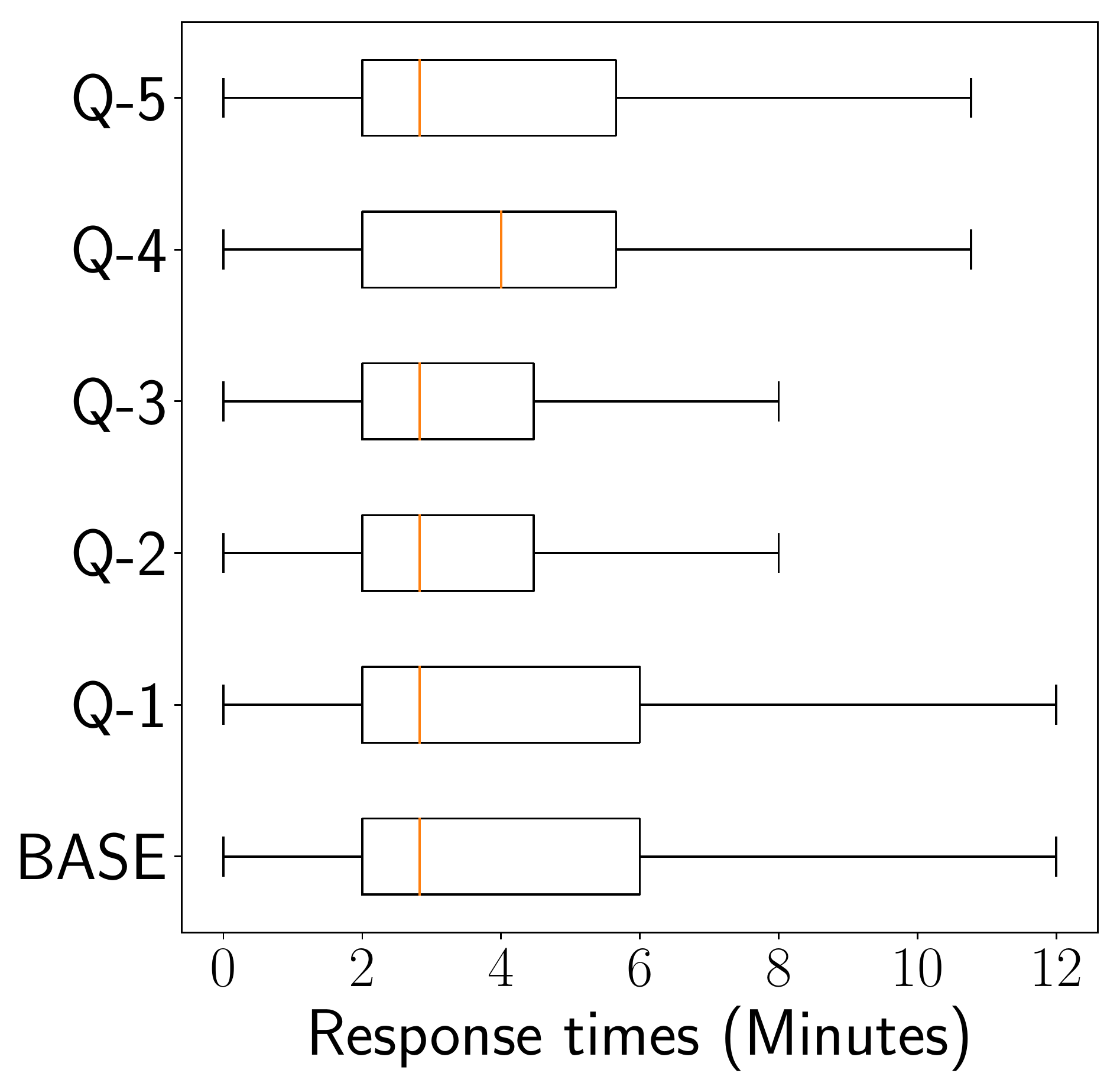}
    \caption{}
    \label{fig:policy_resp}
\end{subfigure}
\begin{subfigure}{.163\linewidth}
  \centering
\includegraphics[width=\linewidth]{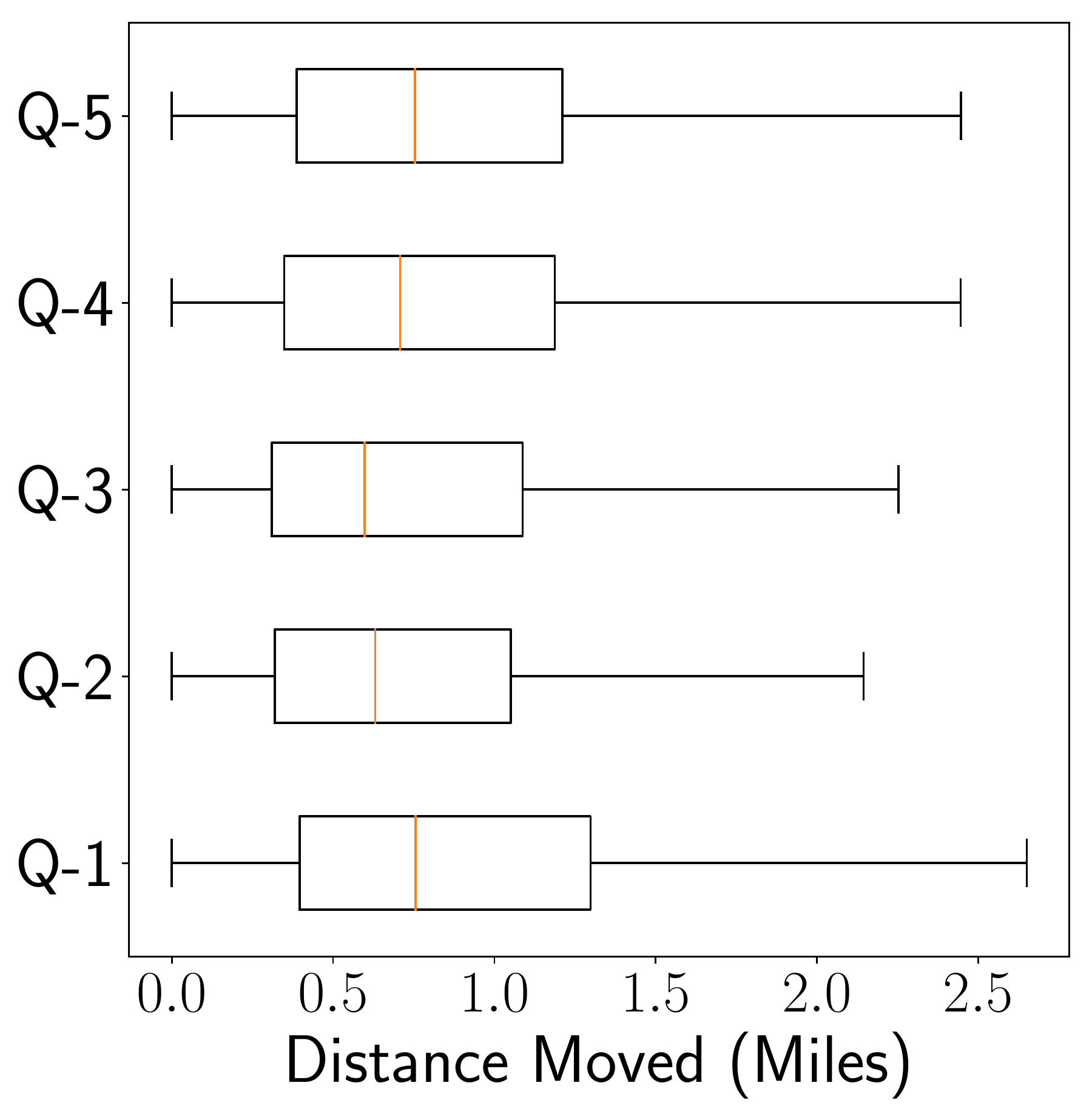}
    \caption{}
    \label{fig:policy_dist}
\end{subfigure}
\begin{subfigure}{.163\linewidth}
  \centering
\includegraphics[width=\linewidth]{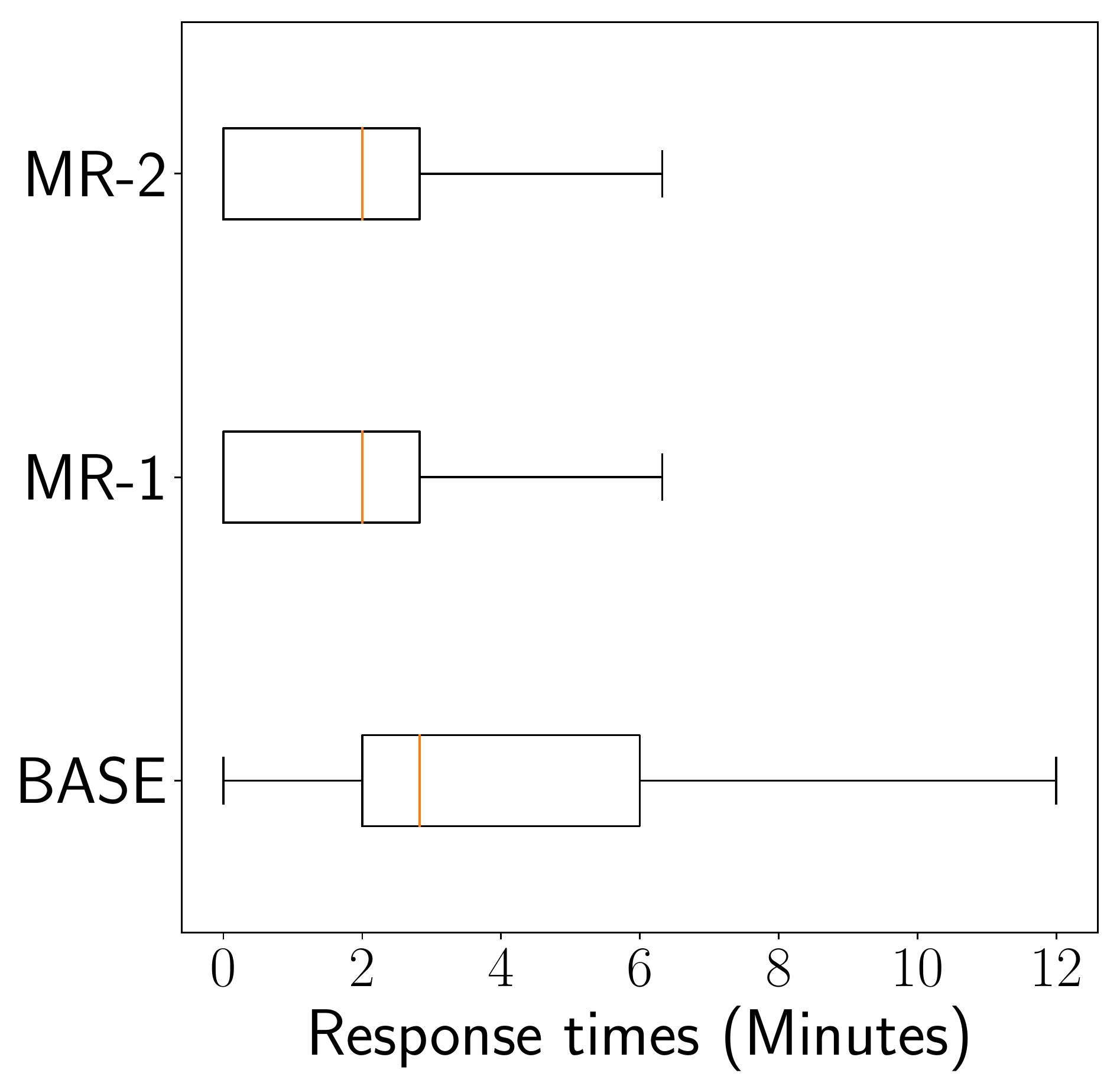}
    \caption{}
    \label{fig:mmcts_oracle_resp}
\end{subfigure}
\begin{subfigure}{.163\linewidth}
  \centering
\includegraphics[width=\linewidth]{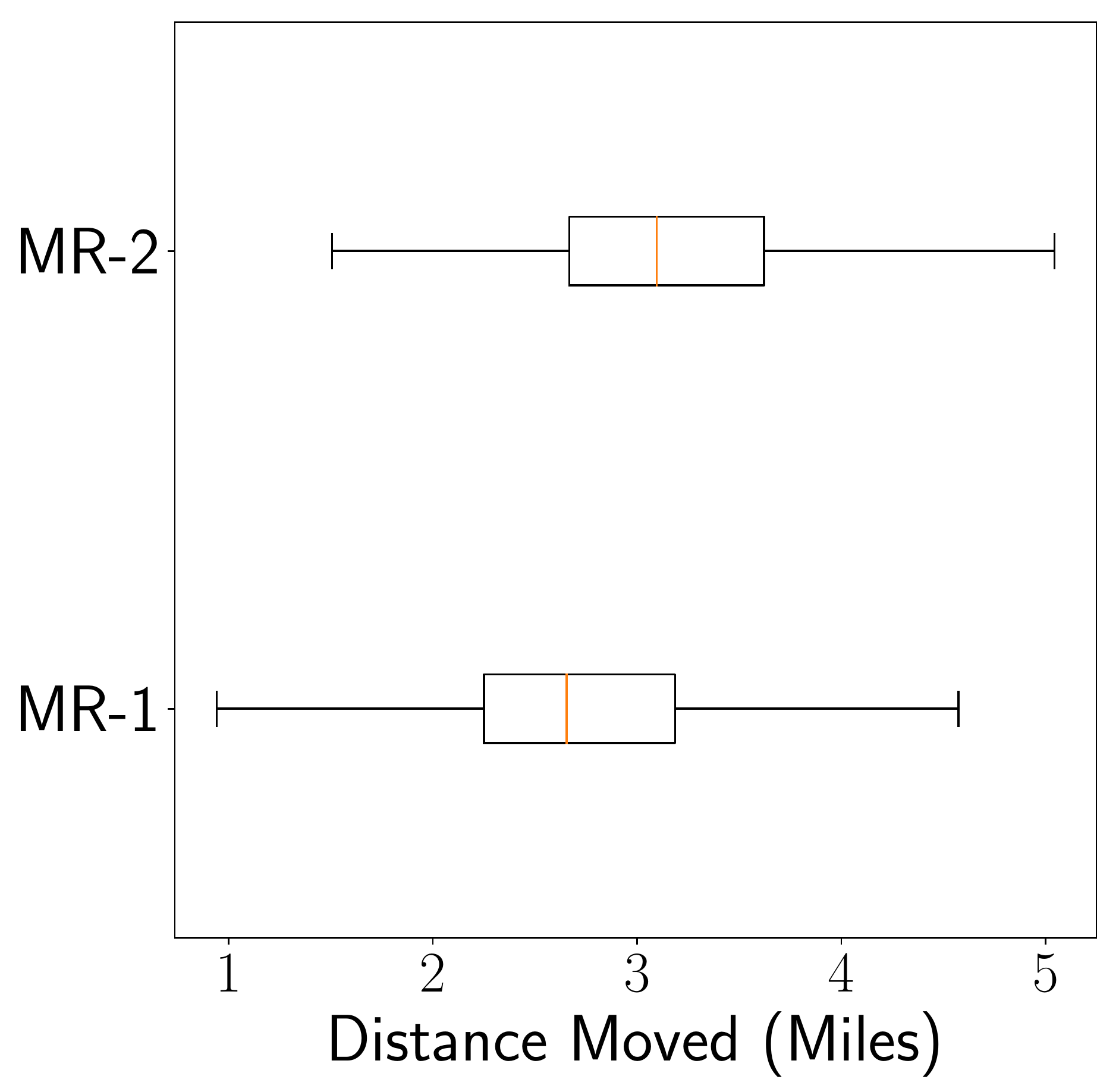}
    \caption{}
       \label{fig:mmcts_oracle_dist}
\end{subfigure}
\begin{subfigure}{.163\linewidth}
  \centering
\includegraphics[width=\linewidth]{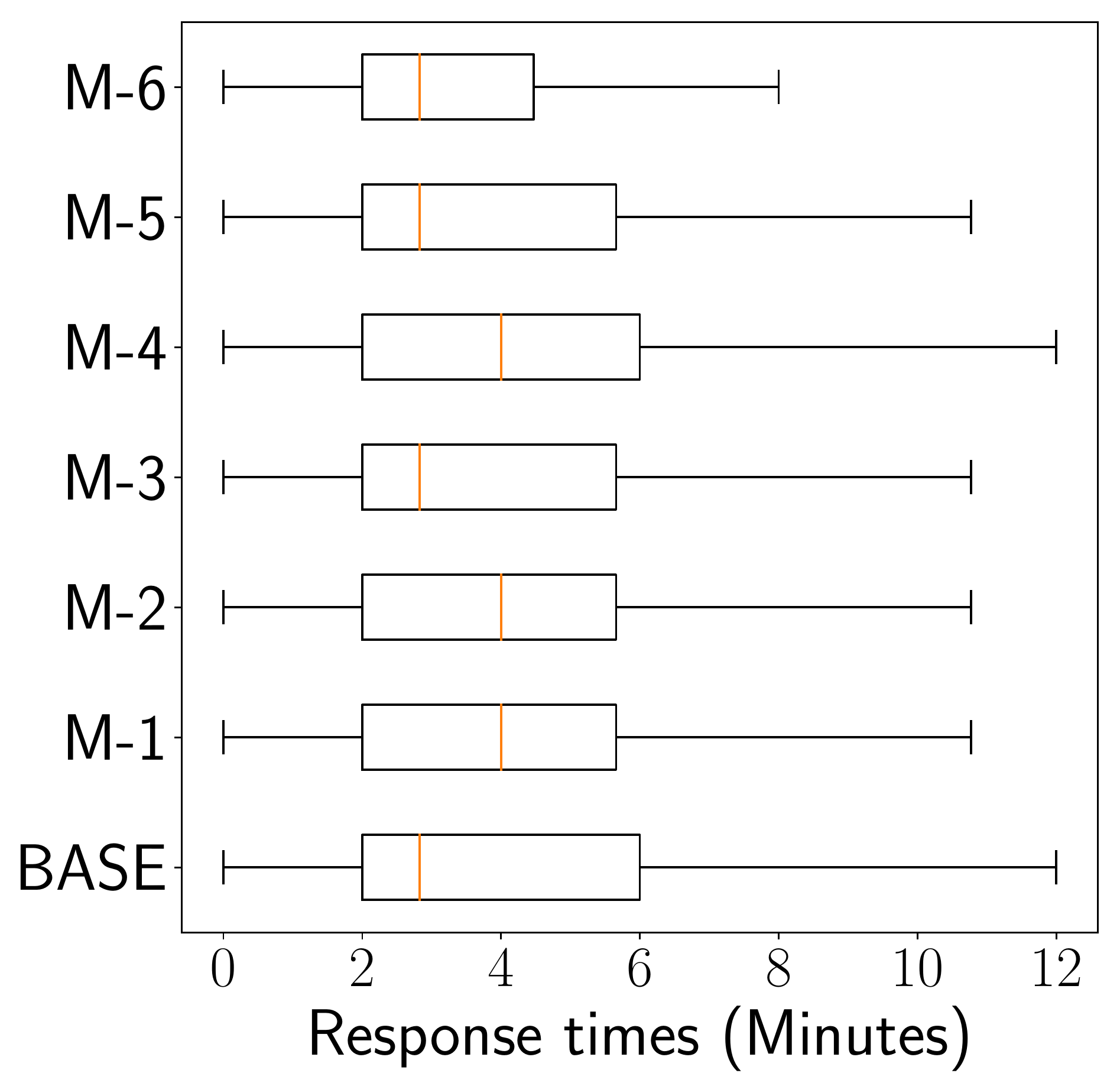}
    \caption{}
   \label{fig:mmcts_search_resp}
\end{subfigure}
\begin{subfigure}{.163\linewidth}
  \centering
\includegraphics[width=\linewidth]{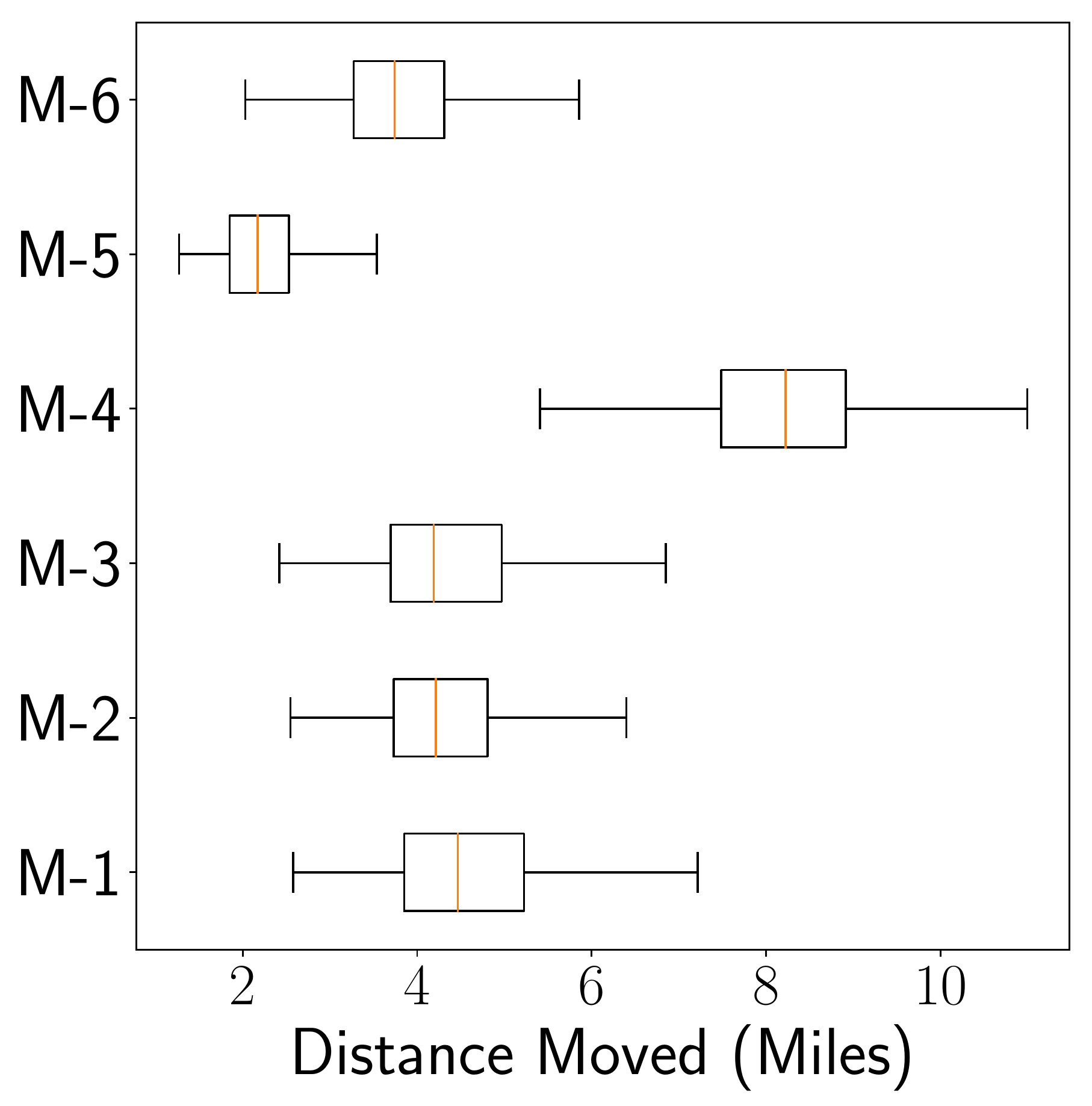}
    \caption{}
       \label{fig:mmcts_search_dist}
\end{subfigure}
\caption{a) The response time distributions for each queue rebalancing policy experiment. b) Distribution of average miles traveled by each responder at each balancing step in the queue rebalancing policy experiments. The baseline approach has no rebalancing, so it is excluded. c) The response time distributions for each MMCTS experiment using an oracle. d) Distributions of average miles traveled by each responder at each balancing step of the MMCTS experiments using an oracle. e) The response time distributions for each MMCTS parameter search experiment. f) Distributions of average miles traveled by each responder at each balancing step of the MMCTS parameter search experiment.}
\label{fig:fig}
\vspace{-0.1in}
\end{figure*}



\section{Results and Discussion}
\label{sec:res_and_discussion}
We now discuss the results of the experiments for the two policies.

\subsection{Queue Based Rebalancing Policy}
We first compare the queue based rebalancing policy described in section \ref{sec:rate_rebal_description} to the baseline policy of no rebalancing. In these experiments rebalancing occurred every half hour, and the incident rates $\upsilon$ were average historical rates from the training data. We tested several values (in miles) for the depots' RoI, and compared the distributions of response times (figure \ref{fig:policy_resp}) and the rebalancing distance traveled by each responder (figure \ref{fig:policy_dist}). 

Our first observation is that increasing the RoI does not necessarily increase performance; there is an optimal zone around RoI=3, implying that encouraging responders to spread out is beneficial. We also see that while Q-3's median and 1st quartile response times remained fairly consistent with the baseline, the upper quartiles are reduced. This decreases the response time's mean and variance, making the system more fair to all incidents. 
We also observe that Q-2 and Q-3's responders traveled less than 1 mile on average each balancing step.   





\subsection{MMCTS Rebalancing}
To determine the potential of the MMCTS rebalancing approach, we first compare the two agent action models described in section \ref{sec:other_agents_models} (\textit{Static Agent Policy} and \textit{Queue Rebalancing Policy}) using \textbf{an oracle}, which has complete information regarding future incidents (this assumption takes the errors of the prediction model out of comparison and enables us to observe the best results that we can obtain) . We present the results for the response time distributions in figure \ref{fig:mmcts_oracle_resp} and the average responder distance traveled per rebalancing step in figure \ref{fig:mmcts_oracle_dist}. 

Our first observation is that the MMCTS approach has high potential. Using an oracle, it is able to significantly decrease the response time distribution compared to the queue based policy above. This is not surprising given that a standard MCTS algorithm given perfect information should perform well given adequate time, but it demonstrates that the MMCTS extensions of independent action evaluation for each agent and action filtering are valid. 
Secondly, we see that MR-1 (using a static agent policy) outperforms MR-2 (using the queue rebalancing policy). 
Last, we observe that responders traveled between 2 and 4 miles on average each during balancing step in these experiments, which is significantly higher than the queuing approach. 

Next, we examine a more the practical approach using the \textbf{incident prediction model} based on survival analysis. Since the static agent policy performed better in the oracle experiments, we use it for these experiments. There are several hyper-parameters that can affect the performance of the algorithm, including 
\begin{enumerate*}
    \item MCTS Iteration Limit 
    \item Rebalancing Period - the amount of time between rebalancing steps
    \item Distance Weight in Reward Function $\psi$ - this represents the importance of distance traveled for rewards
    \item Look-ahead Horizon for MCTS.
\end{enumerate*}

We vary these parameters to see their effect on the system (see table \ref{tab:experimental_table}). We present the response time distributions of MMCTS using the incident model in figure \ref{fig:mmcts_search_resp}, and the average responder distance traveled per rebalancing step in figure \ref{fig:mmcts_search_dist}. 
We observe that different parameter choices lead to different performance characteristics. For example, we see that changing the distance weight has a large impact on the distance responders travel; users with tight budgets for responder movement and maintenance will want to pay close attention to this parameter. 
Comparing the queue based policy with MMCTS, we see that both improve the response time distributions compared to the baseline. MMCTS is more configurable, but is also more sensitive to poor hyper-parameter choices. With proper hyper-parameter choices, both fulfil the constraints discussed in section \ref{sec:problem_def} by having quick dispatching decisions, allowing for limited communication, and allowing users to control for distance traveled (i.e. wear and tear).
\begin{table}[t]
\captionsetup{font=small}
\caption{Outline of the experimental runs performed and their corresponding hyper-parameter choices. ($^{*}$When not indicated, parameters are set to values of M-1, the MMCTS Baseline in the table.)}
\resizebox{\columnwidth}{!}{%
\begin{tabular}{|l|l|l|}
\hline
\multicolumn{1}{|c|}{\textbf{Identifier}} & \multicolumn{1}{c|}{\textbf{Description}}                                                                                                                                                                                                                              & \textbf{Hyper-Parameter Choices}                                                                                                                                                              \\ \hline
BASE                                      & Greedy Baseline Without Rebalancing                                                                                                                                                                                                                                    & N/A                                                                                                                                                                                           \\ \hline
Q-1                                       & Queue Based Rebalancing Policy with RoI of 1                                                                                                                                                                                                                           & RoI = 1                                                                                                                                                                                       \\ \hline
Q-2                                       & Queue Based Rebalancing Policy with RoI of 2                                                                                                                                                                                                                           & RoI = 2                                                                                                                                                                                       \\ \hline
Q-3                                       & Queue Based Rebalancing Policy with RoI of 3                                                                                                                                                                                                                           & RoI = 3                                                                                                                                                                                       \\ \hline
Q-4                                       & Queue Based Rebalancing Policy with RoI of 4                                                                                                                                                                                                                           & RoI = 4                                                                                                                                                                                       \\ \hline
Q-5                                       & Queue Based Rebalancing Policy with RoI of 5                                                                                                                                                                                                                           & RoI = 5                                                                                                                                                                                       \\ \hline
MR-1                                      & \begin{tabular}[c]{@{}l@{}}MMCTS - using an oracle for future incidents\\ and a Static Agent Policy\end{tabular}                                                                                                                                                       & \textit{Same as MMCTS Baseline M-1}                                                                                                                                                           \\ \hline
MR-2                                      & \begin{tabular}[c]{@{}l@{}}MMCTS - using an oracle for future incidents\\ and a Queue Rebalancing Policy\end{tabular}                                                                                                                                                        & \textit{Same as MMCTS Baseline M-1}                                                                                                                                                           \\ \hline
M-1                                       & \begin{tabular}[c]{@{}l@{}}MMCTS - Baseline \\ The foundation for the parameter search.\\ Each parameter varies independently while \\ other parameters retain these values. \\ (All M-* experiments use generated incident \\ chains and a Static Agent Policy)\end{tabular} & \begin{tabular}[c]{@{}l@{}}MCTS Iteration Limit = 250\\ Lookahead Horizon = 120 min\\ Reward Distance Weight $\psi$ = 10\\ Reward Discount Factor = 0.99995\\ Rebalance Period = 60 min\end{tabular} \\ \hline
M-2                                       & MMCTS - Iteration Limit of 100                                                                                                                                                                                                                                         & MCTS Iteration Limit = 100*                                                                                                                                                                   \\ \hline
M-3                                       & MMCTS - Iteration Limit of 500                                                                                                                                                                                                                                         & MCTS Iteration Limit = 500*                                                                                                                                                                   \\ \hline
M-4                                       & MMCTS - Reward Distance Weight $\psi$ of 0                                                                                                                                                                                                                                    & Reward Distance Weight $\psi$ = 0*                                                                                                                                                                   \\ \hline
M-5                                       & MMCTS - Reward Distance Weight $\psi$ of 100                                                                                                                                                                                                                                  & Reward Distance Weight $\psi$ = 100*                                                                                                                                                                 \\ \hline
M-6                                       & \begin{tabular}[c]{@{}l@{}}MMCTS  - Rebalance Period of 30 minutes;  \\ Lookahead Horizon of 30 minutes\end{tabular}                                                                                                                                                   & \begin{tabular}[c]{@{}l@{}}Lookahead Horizon = 30 min\\ Rebalance Period = 30min*\end{tabular}                                                                                                \\ \hline
\end{tabular}%
}
\label{tab:experimental_table}
\vspace{-0.25in}
\end{table}









\section{Conclusion}
Principled approaches to Emergency Response Management (ERM) decision making have been explored, but have failed to be implemented into real systems. We have identified that a key issue with these approaches is that they focus on post-incident decision making. We argue that due to fairness constraints, planning should occur between incidents. We define a decision theoretic model for such planning, and implement both a heuristic search using queuing theory and a Multi Agent Monte Carlo Tree Search planner. We find that these approaches maintain system fairness while decreasing the average response time to incidents.

While the focus of this work is in the ERM domain, there are important takeaways for general agent-based systems: (1) Planning performance is dependent on the quality of the underlying event prediction models. (2) It is imperative to understand the needs and constraints for a target domain when designing a planning approach for it to be accepted in practice. (3) The computational capacity of ``agents'' has evolved in recent decades, and should be used to create decentralized planning approaches. Given these takeaways, we will explore the applicability of this framework to other domains where planning occurs over a spatial-temporally evolving process.

\textbf{Acknowledgement:}
This work is sponsored by the National Science Foundation ( CNS-1640624,  IIS-1814958,  IIS-1905558), the Tennessee Department of Transportation and the Center for Automotive Research at Stanford (CARS). We thank our partners from Metro Nashville Fire Department,  and Metro Nashville Information Technology Services in this work. We would also like to thank Hendrik Baier (CWI) for insights and helpful discussions regarding the paper. We also thank Sayyed Mohsen Vazirizade (Vanderbilt) for his help in reviewing the manuscript and providing important feedback.

\bibliographystyle{ACM-Reference-Format}  
\bibliography{references}  


\end{document}